%% file: main.tex
\documentclass[10pt,twocolumn,letterpaper]{article}

\usepackage[pagenumbers]{cvpr} % To force page numbers, e.g. for an arXiv version


\definecolor{cvprblue}{rgb}{0.21,0.49,0.74}
\usepackage[pagebackref,breaklinks,colorlinks,allcolors=cvprblue]{hyperref}
\usepackage{multirow}

\def\paperID{2119} % *** Enter the Paper ID here
\def\confName{CVPR}
\def\confYear{2025}
\def\sysname{Splatter-360}

\newcommand{\tablestyle}[2]{\setlength{\tabcolsep}{#1}
                            \renewcommand{\arraystretch}{#2}
                            \centering
                            \footnotesize}
\title{Splatter-360: Generalizable 360$^{\circ}$ Gaussian Splatting for Wide-baseline Panoramic Images}

\author{Zheng Chen$^{1}$\thanks{Denotes equal contribution.}~\thanks{This work was done when Zheng Chen interned in Baidu VIS. E-mail: zhengchenecho@gmail.com}~, Chenming Wu$^{2}$\footnotemark[1]~, Zhelun Shen$^{2}$, Chen Zhao$^{2}$, Weicai Ye$^{3}$,
\and
Haocheng Feng$^{2}$, Errui Ding$^{2}$, Song-Hai Zhang$^{1}$\thanks{Corresponding author. E-mail: shz@tsinghua.edu.cn} \vspace{6pt}\\
    $^1$Tsinghua University\\
    $^2$Baidu VIS \\
    $^3$Zhejiang University \\
}

\begin{document}

\maketitle

% \twocolumn[{%
% \renewcommand\twocolumn[1][]{#1}%
% \maketitle

% \begin{center}
%     \vspace{-20pt}
%     \centering
%     \captionsetup{type=figure}
    % \includegraphics[width=\textwidth,height=5cm]{example-image}
    % \captionof{figure}{Test caption}
% \end{center}%

% }]

\input{sec/0_abstract}

\input{sec/1_intro}

\input{sec/2_formatting}

\input{sec/3_finalcopy}

{
    \small
    \bibliographystyle{ieeenat_fullname}
    \bibliography{main}
}

\input{sec/X_suppl}

\end{document}

%% file: sec/0_abstract.tex
\begin{abstract}
% The ABSTRACT is to be in fully justified italicized text, at the top of the left-hand column, below the author and affiliation information.
% Use the word ``Abstract'' as the title, in 12-point Times, boldface type, centered relative to the column, initially capitalized.
% The abstract is to be in 10-point, single-spaced type.
% Leave two blank lines after the Abstract, then begin the main text.
% Look at previous \confName abstracts to get a feel for style and length.

%为了实现沉浸式体验，使用户能够探索虚拟环境六自由度（6DoF）对于各种应用至关重要，例如，虚拟现实（VR）。宽基线全景图通常用于这些场景应用程序，以减少网络带宽和存储需求。然而，
%从这些全景图中合成新颖的视图仍然是一个关键挑战。尽管现有的神经辐射场方法可以产生逼真的视图.在窄基线和密集图像捕获下，由于难以处理宽基线全景图，它们往往会过拟合训练视图
%从稀疏的360°视图中学习精确的几何。为了解决这个问题，我们
%提出宽基线广义球面辐射场PanoGRF
%全景图，构建了包含360°的球形辐射场
%场景先验。与在透视图像上训练的广义辐射场不同，PanoGRF
%避免了全景到透视转换和直接转换的信息丢失
%聚合每个三维采样点的几何形状和外观特征
%基于球面投影的全景视图。此外，由于某些地区
%全景图只能从一个视图看到，而在广角下从其他视图看不见
%在基线设置中，PanoGRF将360°单眼深度先验整合到球面深度估计中，以改善几何特征。实验结果
%多个全景数据集表明，PanoGRF的表现明显优于
%宽基线全景图的最新可推广视图合成方法
%（例如，PanoGRF）和透视图像（例如，MVSplat）。项目页面：
% https://thucz.github.io/PanoGRF/.

% 宽基线全景图通常用于VR、仿真渲染等应用中，以减少网络带宽和存储要求。然而，从这些全景图中合成新观点仍然是一个关键挑战。尽管现有的3DGS方法可以在窄基线和密集图像捕获下产生逼真的视图，但在处理宽基线全景图时，由于难以从稀疏的360°视图中学习精确的几何图形，它们往往会过度拟合训练视图。为了解决这个问题，我们提出了\sysname{}，面向宽基线全景图的可泛化360 Gaussian Splatting，它构建了包含360°场景先验的球面辐射场。与在透视图像上训练的可推广3DGS不同，\sysname{}避免了全景到透视转换的信息丢失，我们基于球形扫描算法构建了球形cost volume来提升网络的深度感知能力,然后基于球形cost-volume预测球形深度和3DGS的属性，然后利用球形投影，将这些3D gaussian primitives反投影到3D空间中。此外，我们构建了多视图感知的attention和mamba2融合的网络作为我们的提取特征的网络，在高分辨率利用mamba2进行长距离依赖学习，在低分辨率我们利用attention网络进行高层次的特征融合，在HM3D和Replica等全景数据集上的实验结果表明，\sysname{}在宽基线全景图方面明显优于最先进的可推广视图合成方法（例如，PanoGRF）和透视图像（例如，MVSplat）。

Wide-baseline panoramic images are frequently used in applications like VR and simulations to minimize capturing labor costs and storage needs. However, synthesizing novel views from these panoramic images in real time remains a significant challenge, especially due to panoramic imagery's high resolution and inherent distortions. Although existing 3D Gaussian splatting (3DGS) methods can produce photo-realistic views under narrow baselines, they often overfit the training views when dealing with wide-baseline panoramic images due to the difficulty in learning precise geometry from sparse 360$^{\circ}$ views. 
This paper presents \textit{Splatter-360}, a novel end-to-end generalizable 3DGS framework designed to handle wide-baseline panoramic images. Unlike previous approaches, \textit{Splatter-360} performs multi-view matching directly in the spherical domain by constructing a spherical cost volume through a spherical sweep algorithm, enhancing the network's depth perception and geometry estimation. Additionally, we introduce a 3D-aware bi-projection encoder to mitigate the distortions inherent in panoramic images and integrate cross-view attention to improve feature interactions across multiple viewpoints. This enables robust 3D-aware feature representations and real-time rendering capabilities. Experimental results on the HM3D~\cite{hm3d} and Replica~\cite{replica} demonstrate that \textit{Splatter-360} significantly outperforms state-of-the-art NeRF and 3DGS methods (e.g., PanoGRF, MVSplat, DepthSplat, and HiSplat) in both synthesis quality and generalization performance for wide-baseline panoramic images. Code and trained models are available at \url{https://3d-aigc.github.io/Splatter-360/}.

%This paper presents \sysname{}, a generalizable 360$^{\circ}$ Gaussian Splatting for wide-baseline panoramic images. Unlike generalizable 3DGS models trained on perspective images, \sysname{} performs multi-view matching directly in the spherical domain by constructing a spherical cost volume using a spherical sweep algorithm, thereby enhancing the network's depth perception capabilities. Based on the spherical cost volume, we predict both spherical depth and 3DGS properties, followed by spherical projection to back-project the 3D Gaussian primitives into world space. We introduce a 3D-aware bi-projection encoder to address the spherical distortion. Experimental results on panoramic datasets such as HM3D and Replica demonstrate that \sysname{} significantly surpasses state-of-the-art feed-forward NeRF models (e.g., PanoGRF) and feed-forward GS methods (e.g., MVSplat, DepthSplat, HiSplat) in handling wide-baseline panoramic images.

\end{abstract}

%% file: sec/1_intro.tex
\section{Introduction}
\label{sec:intro}
In recent years, 360-degree cameras and VR/AR headsets have gained widespread popularity, enabling seamless scene capture and delivering immersive experiences by presenting visuals directly to users. However, due to the labor-intensive nature of data acquisition and the high storage cost, panoramic images in industry settings often exhibit wide baselines. Generating novel views from these wide-baseline panoramas is essential to provide users full freedom of movement within virtual environments.

The emergence of neural radiance fields (NeRF) has demonstrated impressive performance in synthesizing novel views from perspective images. However, NeRF typically requires dense images captured from various angles and positions to address the well-known shape-radiance ambiguity. To reduce the need for dense input, generalizable NeRF appears~\cite{pixelnerf, wang2021ibrnet, trevithick2021grf}. Building on the success of these approaches, previous work such as PanoGRF~\cite{panogrf} aims to mitigate overfitting in spherical radiance fields by leveraging 360-degree scene priors. Despite their strengths, NeRF-based methods rely on implicit representations, and the rendering process demands extensive network evaluations. Consequently, these methods require powerful GPUs, making them unsuitable for real-time applications on lightweight mobile devices like VR headsets, where low-power processing is essential.
In contrast to NeRF, 3D Gaussian Splatting (3DGS)~\cite{kerbl20233d} moves away from implicit representation, instead adopting an explicit ellipsoid primitive representation. This representation is particularly well-suited for real-time rendering on traditional rasterization-based devices. However, like NeRF, 3DGS faces challenges with overfitting training views, especially when handling wide-baseline panoramic images.

This paper addresses the challenges of learning from wide-baseline panoramic images for real-time novel view rendering of 3DGS. Unlike previous generalizable 3DGS methods such as~\cite{pixelsplat,mvsplat,li2024ggrt}, our approach, dubbed as \textit{Splatter-360}, operates end-to-end with panoramic image inputs and outputs. The design of our method is motivated by the following key factors:1) Panoramic images provide a full 360$^\circ$ field-of-view (FoV), whereas cubemap perspective images split the scene into six independent views, causing the issue of projecting points behind the camera in the source view when sampling point in the planar cost volume of the reference view — an issue that panoramic images naturally circumvent; 2) Our method eliminates the explicit back-and-forth conversion between panoramic and cube map representations, where backward conversions often introduce seam artifacts during stitching~\cite{bai2024360}.

% for each 3D sampling point in the planar cost volume of the reference view, it is possible that these points may project behind the camera in the source view (i.e., $z$-depth $< 0$). When this occurs, incorrect local image features from the source view are sampled, leading to inaccurate local similarity computations

The end-to-end approach for generalizable 3D Gaussian Splatting (3DGS) presents two major challenges. First, panoramic images have significantly higher resolution than perspective images, requiring an extremely efficient network to avoid excessive memory consumption. To address this, we propose performing multi-view matching directly in the spherical domain by constructing a spherical cost volume through a spherical sweep algorithm. Second, the uneven distortion inherent in panoramic images complicates accurate depth estimation. We introduce a 3D-aware bi-projection encoding that leverages monocular depth, equirectangular projection, and cube-map branch information to tackle this issue. Additionally, by incorporating a cross-view attention mechanism to enhance feature interaction across different viewpoints, we obtain a robust 3D-aware feature representation.

Our contributions can be summarized as follows:
\begin{itemize}
    \item We propose \textit{Splatter-360}, a generalizable 3D Gaussian Splatting method for wide-baseline panoramic images, which is an end-to-end trainable network that can generate 3DGS primitives for novel panoramic view synthesis and enables real-time rendering experience.
    \item We introduce a spherical cost volume based on a spherical sweep algorithm in the feed-forward Gaussian splatting framework, which improves the geometry estimation capabilities of the network.
    \item Extensive experiments conducted on the HM3D~\cite{hm3d} and Replica~\cite{replica} datasets demonstrate that our Splatter-360 significantly outperforms state-of-the-art methods in handling wide-baseline panoramic images, both in terms of synthesis quality and generalization performance.
\end{itemize}

%% file: sec/2_formatting.tex
\section{Related Work}
\label{sec:related_work}

\subsection{Generalizable Novel View Synthesis}
% Novel view synthesis has been a long-standing problem in computer vision, with applications ranging from virtual reality to video compression. Currently, NeRF and 3D Gaussian Splatting (3DGS) has been two mainstreams for perspective view synthesis: (1) \textbf{NeRF} has become widely used for object \cite{nerf_object2, nerf_object1} and scene-level \cite{nerf_scene1, nerf_scene2} perspective view synthesis, in which feed-forward models have gained significant interest due to their faster inference times. (2) \textbf{3D Gaussian Splatting} further enhances the efficiency of novel view synthesis by employing a splatting-based method, which avoids the computationally expensive volume rendering process typically associated with NeRF. Moreover, recent work has proposed a variety of feed-forward 3DGS \cite{mvsplat, depthsplat, charatan2024pixelsplat, lara} models, demonstrating its effectiveness for this task. However, most of these methods are designed for perspective images and struggle with panoramic inputs due to the large field of view and wide baselines. Hence, in this paper, we propose Splatter-360 to extend the strengths of 3DGS to panoramic image.

Recent advancements in novel view synthesis have been largely driven by NeRF and 3D Gaussian Splatting (3DGS). These methods can be broadly classified into two categories: per-scene optimization approaches~\cite{reconfusion, dsnerf, sparf} and generalizable models~\cite{pixelnerf, murf, wide_baseline, pixelsplat, mvsplat, flash3d}. Generalizable methods~\cite{pixelnerf, murf, wide_baseline, pixelsplat, mvsplat, flash3d} enable rapid reconstruction by learning priors from large-scale datasets, allowing for efficient feedforward inference. Below, we provide an overview of generalizable NeRF and 3DGS techniques:

\textbf{Generalizable NeRF:} Early attempts in generalizable NeRF models aim to reconstruct objects and scenes by generating pixel-aligned features for radiance field prediction, pioneered by \cite{pixelnerf, wang2021ibrnet, trevithick2021grf}. Subsequent work such as NeuRay~\cite{liu2022neural} predicts the visibility of 3D points relative to input images, focusing more on visible features. \cite{wide_baseline} and \cite{wang2022attention} further propose a multi-view transformer encoder with epipolar line sampling to capture multi-view geometric priors efficiently. MuRF \cite{murf} uses a target view frustum volume aligned with the target image plane to enable sharp, high-quality rendering. More recently, LRM~\cite{hong2023lrm} and Instant3D~\cite{li2023instant3d} designed a large transformer model to regress triplane NeRF feature for single- or sparse-view reconstruction.

\textbf{Generalizable 3DGS:} To address NeRF’s high computational demands, generalizable 3DGS has emerged to simplify view synthesis through rasterization-based splatting, bypassing the need for expensive volume sampling. Methods like Splatter Image \cite{splatter_image} demonstrate efficient object-level reconstruction by learning Gaussian parameters from single views. In contrast, pixelSplat \cite{pixelsplat} and Flash3D \cite{flash3d} have pushed this technique to scene-level reconstruction. More recently, ~\cite{zhang2025gs} extends the paradigm of LRM~\cite{hong2023lrm} to scene-level with 3DGS, MVSplat~\cite{mvsplat} introduces a cost volume representation using plane sweeping to boost performance further. HiSplat~\cite{hisplat} proposes a hierarchical structure for generalizable 3DGS. DepthSplat~\cite{depthsplat} leverages depth estimation to address multi-view depth methods' limitations and failure cases. However, most of these methods are designed for perspective images and struggle with panoramic inputs due to the large field of view and wide baselines. In response, we propose Splatter-360, extending the strengths of 3DGS to panoramic images.

% \chenming{Fsgs: Real-time few-shot view synthesis using gaussian splatting}

% \chenming{G3R: Gradient Guided Generalizable Reconstruction}

% \chenming{Mvsgaussian: Fast generalizable gaussian splatting reconstruction from multi-view stereo}

% % Generation

% \chenming{ZeroRF: fast sparse view 360 reconstruction with zero pretraining}

% \chenming{Lrm: Large reconstruction model for single image to 3d}

% \chenming{Long-LRM: Long-sequence Large Reconstruction Model for Wide-coverage Gaussian Splats}

% \chenming{GS-LRM: Large Reconstruction Model for 3D Gaussian Splatting}

% \chenming{Flash3D: Feed-Forward Generalisable 3D Scene Reconstruction from a Single Image}

% \chenming{InstantSplat: Sparse-view SfM-free Gaussian Splatting in Seconds}

% \chenming{Dust3r}

\subsection{Panoramic View Synthesis and Generation}
Unlike perspective, novel view synthesis, panoramic novel view synthesis introduces unique challenges due to the distortions caused by equirectangular projection. Early works~\cite{attal2020matryodshka, habtegebrial2022somsi} attempted to synthesize panoramic views using multi-sphere images. More recent approaches have focused on reconstructing radiance fields or 3D Gaussian scenes from dense panoramic inputs~\cite{gu2022omni, bai2024360}. The challenge intensifies with sparse 360$^\circ$ inputs, as depth estimation becomes significantly more difficult.
A few methods seek to address this by leveraging the predicted depth and geometric warping to construct neural radiance fields for panoramic novel view synthesis~\cite{kulkarni2023360fusionnerf, chang2023depth}. 360Roam~\cite{360roam} proposes to adapt neural rendering methods for panoramic inputs, struggling in wide-baseline scenarios. PERF~\cite{perf}, on the other hand, explores an inpainting-based approach to generate neural radiance fields from a single panoramic image.
Most relevant to our work is PanoGRF~\cite{panogrf}, which introduces Generalizable Spherical Radiance Fields for wide-baseline panoramas. PanoGRF achieves state-of-the-art performance by directly aggregating geometry and appearance features of 3D sample points from each panoramic view. However, its inference and rendering speed are limited, making it unsuitable for real-time applications. In contrast, our work proposes a generalizable panoramic view synthesis pipeline designed to generate 3D Gaussian primitives, carefully considering network design and inherent real-time performance during rendering.

In contrast to the generalizable setting, several works have explored text-based panoramic view generation, albeit with distinct objectives. Nonetheless, these approaches often share similar network design philosophies. For example, Text2Light~\cite{chen2022text2light} focuses on high dynamic range panorama generation, while FastScene~\cite{ma2024fastscene} employs coarse view synthesis followed by progressive inpainting-based generation. DiffPano~\cite{diffpano} introduces a spherical epipolar-aware diffusion model for panorama synthesis. Additionally, Panfusion~\cite{zhang2024taming}, Dreamscene360~\cite{zhou2025dreamscene360}, and SceneDreamer360~\cite{li2024scenedreamer360} propose novel architectures built on 2D diffusion models. Although these works are orthogonal to ours, they share several challenges in handling panoramic view synthesis.

\vspace{4pt}

\noindent \textit{Concurrent work.} 
While preparing our work, we found several concurrent works address similar challenges using diffusion priors. For instance, ViewCrafter~\cite{yu2024viewcrafter}, GaussianEnhancer~\cite{liu20243dgs}, ReconX~\cite{liu2024reconx}, and FreeVS~\cite{wang2024freevs} all use video diffusion models to improve the novel view synthesis. However, these works require computationally intensive de-noising and are not designed for panoramic view synthesis. A very recent work, MVSplat-360~\cite{chen2024mvsplat360}, focuses on synthesizing novel perspective images that enable arbitrary position and viewpoint exploration (i.e., 360$^\circ$ navigation) using video diffusion priors, aligning with the goals of aforementioned methods but in generalizable 3DGS setting. In contrast, our work specifically tackles generating novel views for panoramic images (i.e., 360$^\circ$ images).

%% file: sec/3_finalcopy.tex
\section{Proposed Method}
\label{sec:method}

\begin{figure*}
\centering
\includegraphics[width=1.0\textwidth]{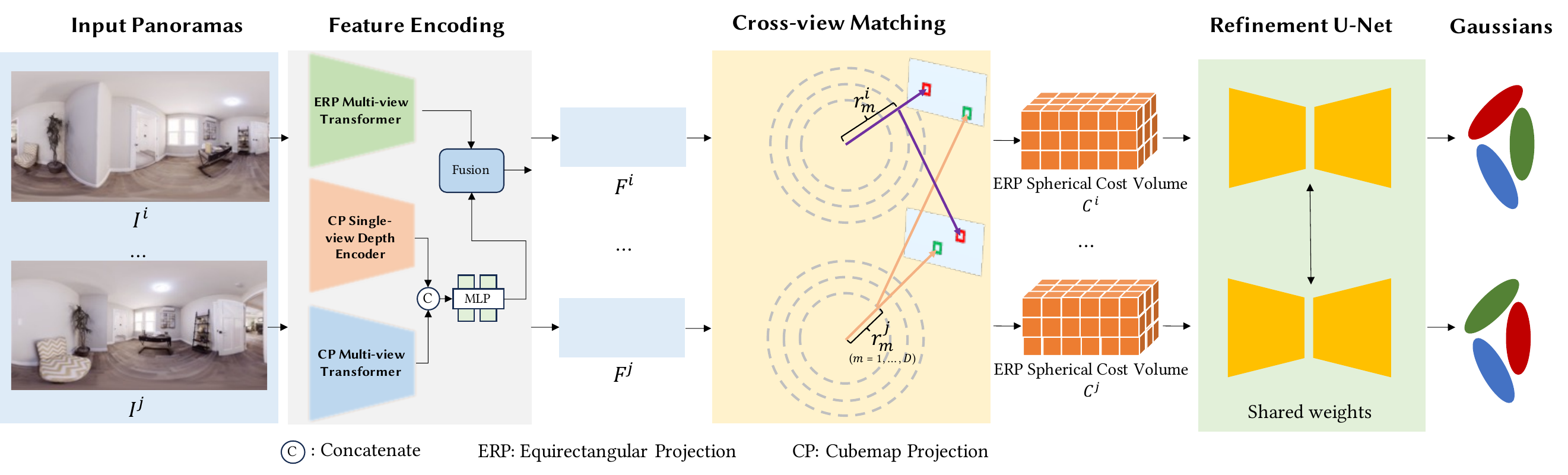}
\caption{Our Splatter-360 processes 360° panoramic images using a bi-projection encoder that extracts features from both equirectangular projection (ERP) and cube-map projection (CP) through multi-view transformers. These features are used for spherical cost volume construction, and multi-view matching is performed between the reference and source views in spherical space. Next, a refinement U-Net is applied to enhance the spherical cost volume, yielding refined cost volumes and more accurate spherical depth estimations. These refined outputs are then fed into the Gaussian decoder, which produces pixel-aligned Gaussian primitives for synthesizing novel views.} 

\label{fig:pipeline}
\end{figure*}

% \subsection{Pipeline of Splatter-360}

Our proposed framework, Splatter-360, is designed to synthesize novel views from wide-baseline 360° panoramic images by leveraging the strengths of 3DGS. 
Unlike conventional 3DGS methods \cite{mvsplat, depthsplat}, which are trained on perspective images, Splatter-360 operates directly on panoramic images, thereby mitigating the information loss associated with perspective transformation. 
The overall pipeline is depicted in Fig.~\ref{fig:pipeline}. A detailed explanation of the 3D-aware bi-projection encoding is provided in Sec.~\ref{sec:3d-biprojection}, spherical depth estimation is discussed in Sec.~\ref{sec:spherical-cv}, and the pixel-aligned Gaussian decoding is outlined in Sec.~\ref{sec:pixel-aligned}. Additionally, preliminary on 3DGS is included in the supplementary material to ensure this paper is self-contained.

% TODO: gaussian Splatting formula.
\subsection{3D-Aware Bi-Projection Encoding}
\label{sec:3d-biprojection}
% 使用ERP映射优点在于视野广，图像连续性好，但简单地使用等矩形映射(ERP)作为输入并不是一个最优的解法：球形扭曲导致图像展开不均匀问题，对于卷积网络来说，难以在这些扭曲的地方难以学习正确的特征。为此我们使用双映射网络提取特征，除了ERP以外，我们额外使用cubemap映射来提取辅助的特征。

Although equirectangular projection (ERP) provides a wide field of view and seamless image continuity, the significant distortions, especially near the poles, hinder the network's ability to learn meaningful features effectively. To address this, we propose a bi-projection encoder that leverages both the advantages of ERP and the complementary strengths of cube-map projection (CP). By extracting auxiliary features through CP, our approach enhances the network's capacity to handle these distortions better.

The overall structure of the bi-projection encoder is illustrated on the left side of Fig.~\ref{fig:pipeline}. In each branch of the bi-projection network, we begin by applying the convolutional neural network from Unimatch~\cite{unimatch} to independently extract local features from both the ERP and CP representations. Unlike prior panorama feature extraction methods~\cite{unifuse, bifuse}, which primarily focus on 2D features, our network is designed to learn 3D-aware features. To achieve this, we employ a cross-view attention mechanism to facilitate feature interactions between different viewpoints in the ERP and across the various views in the CP. We obtained $\boldsymbol{F}_{ERP}$ and $\boldsymbol{F}_{CP}$ after the cross-view attention module. $\boldsymbol{F}_{CP}$ is then stitched up into ERP feature $\boldsymbol{F}_{C2E}$.
% \paragraph{Cross-view Attention}

For wide-baseline panoramas, naive 360$^\circ$ multi-view matching methods are hard to handle occlusions and textureless areas. To enhance the 3D-aware capability of the encoder in reasoning about geometric details, we introduce the CP single-view depth encoder. This is accomplished by incorporating the single-view geometric priors from a pre-trained monocular depth network into the CP branch at the feature level. Specifically, we extract monocular depth features from each perspective view using a pre-trained depth estimation network DepthAnythingV2~\cite{depthanythingv2}. Once the monocular depth feature maps $\boldsymbol{F}^{mono}_{CP}$ for all perspective views are obtained, we stitch these CP features into the ERP view, denoted as $\boldsymbol{F}^{mono}_{C2E}$.
Next, we concatenate $\boldsymbol{F}^{mono}_{C2E}$ with the transformed CP feature $\boldsymbol{F}_{C2E}$, and use a two-layer MLP to get the final CP branch feature:
% For wide-baseline panoramas, naive 360-degree multi-view matching is hard to handle occlusions and textureless area. To enhance the 3D-aware capability of the encoder in reasoning about geometric details, we introduce monocular depth prediction features. This is accomplished by incorporating the single-view geometric priors from a pretrained monocular depth network into the CP encoder at the feature level. Specifically, we extract single-view features for each perspective view using the Vision Transformer (ViT) from DepthAnythingV2~\cite{depthanythingv2}. Once the single-view feature maps $\boldsymbol{F}^{mono}_{CP}$ for all perspective views are obtained, we stitch these CP features into an ERP feature, denoted as $\boldsymbol{F}^{mono}_{C2E}$.
% Next, we concatenate $\boldsymbol{F}^{mono}_{C2E}$ with the native ERP feature $\boldsymbol{F}_{C2E}$, and feed the combined feature into a two-layer MLP, $\mathcal{F}_1$, to compute the fused feature:
\begin{equation}
    \boldsymbol{F}^\prime_{C2E} = \mathcal{F}_1([\boldsymbol{F}^{mono}_{C2E}, \boldsymbol{F}_{C2E}]),
\end{equation}
where $[.,.]$ denotes the concatenation operation and $\mathcal{F}_1$ is a two-layer MLP. Then, we propose a fusion module, $\mathcal{F}_2$, to fuse the CP branch feature $\boldsymbol{F}^\prime_{C2E}$ and ERP branch feature $\boldsymbol{F}_{ERP}$. The final output feature of our encoder is represented as:
\begin{equation}
    \boldsymbol{F} = \mathcal{F}_2(\boldsymbol{F}^\prime_{C2E}, \boldsymbol{F}_{ERP}),
\end{equation}
where $\mathcal{F}_2$ is the proposed fusion module, which consists of convolutional layers augmented with squeeze-and-excitation blocks~\cite{se,unifuse}.

\subsection{Spherical Depth Estimation}
\label{sec:spherical-cv}
% 如果将全景图拆成普通图像，对于参考视角的planar cost-volume的3D采样点来说，它有可能在投影到源视角时，出现在源视角相机的背后(z-depth < 0)，这个时候会采样到错误的源视角局部图像特征，这时候计算得到的局部相似性往往会误导深度的计算。而这个问题的根源在于普通的cost volume使用的窄视野的平面扫描算法，因此，本篇文章利用全景图360视野的特点，采用球形cost volume来预测深度。

% It is straightforward to convert a panorama into cube map-like perspective images for the utilization of perspective-based methods~\cite{mvsplat, depthsplat} for feed-forward 3DGS. However, for each 3D sampling point in the planar cost volume of the reference view, there is a possibility that these points may project behind the camera in the source view (i.e., $z$-depth $< 0$). In such cases, incorrect local image features from the source view are sampled, leading to inaccurate local similarity computations, which in turn misguide the depth estimation. This issue stems from the limitations of the narrow field-of-view plane sweep algorithm employed in standard cost volume construction.
% To address this, we leverage the 360-degree field of view inherent to panoramas and apply the spherical projection formula derived from DiffPano~\cite{diffpano} to construct a spherical cost volume.
After obtaining a robust feature representation for panoramic images, the next step is cost volume construction. One straightforward approach is to convert the panorama into cube-map-like perspective images and then follow the perspective-based methods such as MVSplat \cite{mvsplat} and DepthSplat \cite{depthsplat} to construct cost volume. However, such a method has an unavoidable shortcoming: for each 3D sampling point in the planar cost volume of the reference view, these points may project behind the camera in the source view (i.e., $z$-depth $< 0$). When this occurs, incorrect local image features from the source view are sampled, leading to inaccurate local similarity computations and, ultimately, erroneous depth estimates. This issue is a consequence of the narrow field-of-view plane sweep algorithm traditionally used for the perspective-based method.

To address this, we leverage the 360-degree field of view inherent to panoramas and apply the spherical projection formula derived from \cite{diffpano} to construct a spherical cost volume. More concretely, the spherical sweep algorithm samples depth candidates within the spherical domain, computing feature correlations across multiple views to construct the cost volume. For each reference view, we sample $D$ depth candidates in the spherical space, where the depth values $r_m \in [r_{\text{near}}, r_{\text{far}}]$ are sampled in the logarithmic space between the near range $r_{\text{near}}$ and the far range $r_{\text{far}}$. 

To transform the equirectangular coordinates to spherical coordinates, we use the following equations:

% equirectangular to spherical
\begin{equation}
\label{sphere_cord}
\left\{
\begin{aligned}
\theta &= (0.5 - \frac{u}{W}) \cdot 2\pi \\
\phi &=(0.5 - \frac{v}{H}) \cdot \pi,
\end{aligned}
\right.
\end{equation}
where $u$ and $v$ represent the pixel coordinates in the equirectangular grid, while $\theta$ and $\phi$ denote the longitude and latitude, respectively. Additionally, $W$ and $H$ refer to the width and height of the ERP feature maps on which multi-view matching is performed.

Next, we transform the spherical polar coordinates $(r, \theta, \phi)$ into Cartesian coordinates $(x, y, z)$ to obtain the camera coordinates $\boldsymbol{p}^i_{\text{camera}}$
% spherical to cartesian
\begin{equation}
\label{cam_cord}\left\{
\begin{aligned}
x_{cam} &= r \cos(\phi) \cdot \sin(\theta) \\
y_{cam} &= r \sin(\phi) \\
z_{cam} &= r \cos(\phi) \cdot \cos(\theta),
\end{aligned}\right.
\end{equation}

To obtain the pixel coordinates $(u,v)$ at the source view $j$, we back-project the camera coordinates using the relative camera pose $W^{i\rightarrow j}$ between the reference view and the source view:
\[
\boldsymbol{p}^j_{camera} = W^{i\rightarrow j}\boldsymbol{p}^i_{camera}
\]
With known corresponding pixel coordinates, we sample the local image feature $F^{j\rightarrow i}$ from the source view feature and compute the feature similarity between $F^{j\rightarrow i}$ and $F^i$ as follows: 
\begin{equation}
\label{eq:correlation}
    {\boldsymbol{C}}_{r_m}^{i} = \frac{{\boldsymbol{F}}^i \cdot {\boldsymbol{F}}^{j \to i}_{r_m}}{\sqrt{C}} \in \mathbb{R}^{H \times W}, \quad m = 1, 2, \cdots, D,
\end{equation}
where $C$ denotes the total feature channel and ${C}_{r_m}^{i}$ denotes the feature similarity between $F^{j\rightarrow i}$ and $F^i$ at the $m^{th}$ depth candidate. Then, we can concatenate each depth candidates' feature similarity to get the spherical cost volume. 
\begin{equation}
\label{eq:cost_volume}
    \boldsymbol{C}^i = [\boldsymbol{C}_{r_1}^{i}, \boldsymbol{C}_{r_2}^{i}, \cdots, \boldsymbol{C}_{r_D}^{i}] \in \mathbb{R}^{H \times W \times D}.
\end{equation}
As the initial spherical cost volume only considers the pixel-level similarity and is hard to handle occlusions and textureless areas. We further use a U-Net \cite{mvsplat} to refine it. Specifically, the proposed U-Net mainly learns a residual volume value $ \Delta \boldsymbol{C}^{i}$, which can be added to the initial spherical cost volume to get the final one:
% To obtain a residual volume value $ \Delta \boldsymbol{C}^{i}$, we use the U-net from Stable Diffusion~\cite{sd} similar to~\cite{mvsplat}, and add $ \Delta \boldsymbol{C}^{i}$ to the naive spherical cost volume
% $\Delta \boldsymbol{C}^{i}$:

\begin{figure*}
\centering
\includegraphics[width=\textwidth]{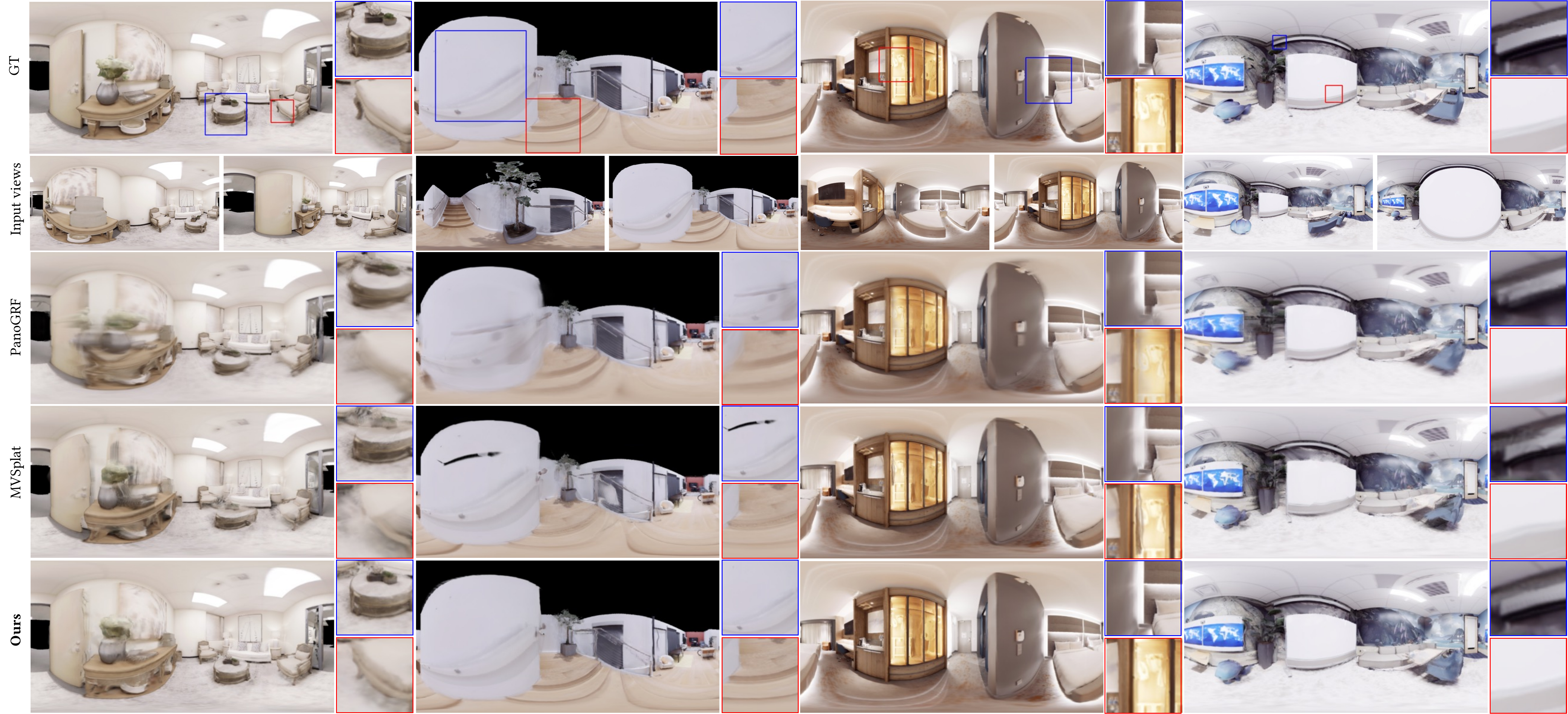}
\caption{Qualitative comparison between our Splatter-360 and PanoGRF, MVSplat on the Replica dataset. Regions with notable differences are highlighted using red and blue rectangles. Please zoom in for a clearer view.}
\label{fig:replica_cmp}
% \vspace{-2pt}
\end{figure*}
\begin{equation}
    \tilde{\boldsymbol{C}}^{i} = \boldsymbol{C}^i + \Delta \boldsymbol{C} ^i \in \mathbb{R}^{H \times W \times D}.
\end{equation}
% TODO:参考MVSplat的特征相似性计算过程，只不过warping过程发生了变化：
% \paragraph{spherical depth estimation} 
Finally, we can use the softmax operation to get the final spherical depth estimation:
\begin{equation}
\label{eq:softmax_depth}    
    \boldsymbol{D}^i = \mathrm{softmax} (\tilde{\boldsymbol{C}}^i)\boldsymbol{G} \in \mathbb{R}^{H \times W},
\end{equation}
where $\boldsymbol{G} = [r_1, r_2, \dots, r_D] \in \mathbb{R}^{D}$ is the predefined depth candidates and $softmax$ denotes the softmax function which can transform spherical cost volume to the probability distribution for each depth candidates.

% One of the key components of Splatter-360 is the spherical cost volume, which is constructed using a spherical sweep algorithm. This cost volume allows the network to perceive depth across the entire 360° field of view, improving the accuracy of geometry estimation for wide-baseline panoramic images. 

\subsection{Pixel-Aligned Gaussian in Equirect-Coordinate}
\label{sec:pixel-aligned}
%360$^{\circ}$ Gaussian Parameters }
% Once the essential features and spherical cost volume have been constructed, we proceed to the decoding phase of the pipeline. In this phase, we predict pixel-aligned Gaussians on the equirectangular coordinate grid, where each pixel generates a corresponding Gaussian primitive. Specifically, for each pixel, we estimate Gaussian centers, quaternions, spherical harmonics (SH), and scaling factors, respectively.
Once the 3D-aware feature representation and spherical cost volume are obtained, we proceed to predict pixel-aligned Gaussians on the equirectangular coordinate grid. Specifically, for each pixel, we estimate Gaussian centers, quaternions, spherical harmonics (SH), and scaling factors, respectively.

\vspace{4pt} \noindent \textbf{Gaussian centers $\boldsymbol{\mu}$.} We use the estimated spherical depth $\boldsymbol{D}$ to calculate the gaussian centers. Specifically, for each pixel, we first compute the camera cartesian coordinates $\boldsymbol{p}_{\text{camera}}$ of the gaussian center using its spherical depth $\boldsymbol{D}$ via Eq.~\ref{sphere_cord} and Eq.~\ref{cam_cord}. Next, the world coordinates of the Gaussian center are calculated as $\boldsymbol{p}_{\text{world}} = W \boldsymbol{p}_{\text{camera}}$, where $W$ represents the camera-to-world transformation matrix. 
%\zhelun{please highlight the importance of spherical depth in this stage}

\vspace{4pt} \noindent \textbf{Opacity $\alpha$.} The opacity is calculated according to the matching confidence. Specifically, we first use the probability distribution of spherical cost volume(i.e., the softmax output of Eq. \ref{eq:softmax_depth}) to get the matching confidence. Then, we feed the matching confidence along with the image features into a two-layer convolutional network to predict the opacity.

\vspace{4pt} \noindent \textbf{Covariance $\boldsymbol{\Sigma}$ and SH $\boldsymbol{c}$.}
% TODO:和MVSplat一样
% \chenzheng{similar to MVSplat}
We predict these parameters using two convolutional layers, which take as input the concatenation of image features, the refined cost volume, and the original ERP images. The covariance matrix $\Sigma$ is computed as:
\[
\Sigma = R(\theta)^\top \, \text{diag}(s) \, R(\theta),
\]
following the formulation in~\cite{pixelsplat}. Here, $R(\theta)$ is the rotation matrix parameterized by quaternions, $s$ represents the scaling matrix, and $\boldsymbol{c}$ is the spherical harmonics for direction-dependent color encoding.

%------------------------------------------------------------------------
\section{Experiments}
% \begin{figure*}[ht]
%     \begin{center}
%         \centerline{\includegraphics[width=1\linewidth]{pics/replica_cmp.png}}
%         \vspace{-0.1cm}
%         \caption{{Novel view synthesis comparison between PanoGRF, MVSplat and \sysname{}.}}
%         \label{fig:replica_cmp}
%     \end{center}
%     \vspace{-0.6cm}
% \end{figure*}

\begin{figure*}
\centering
\includegraphics[width=\textwidth]{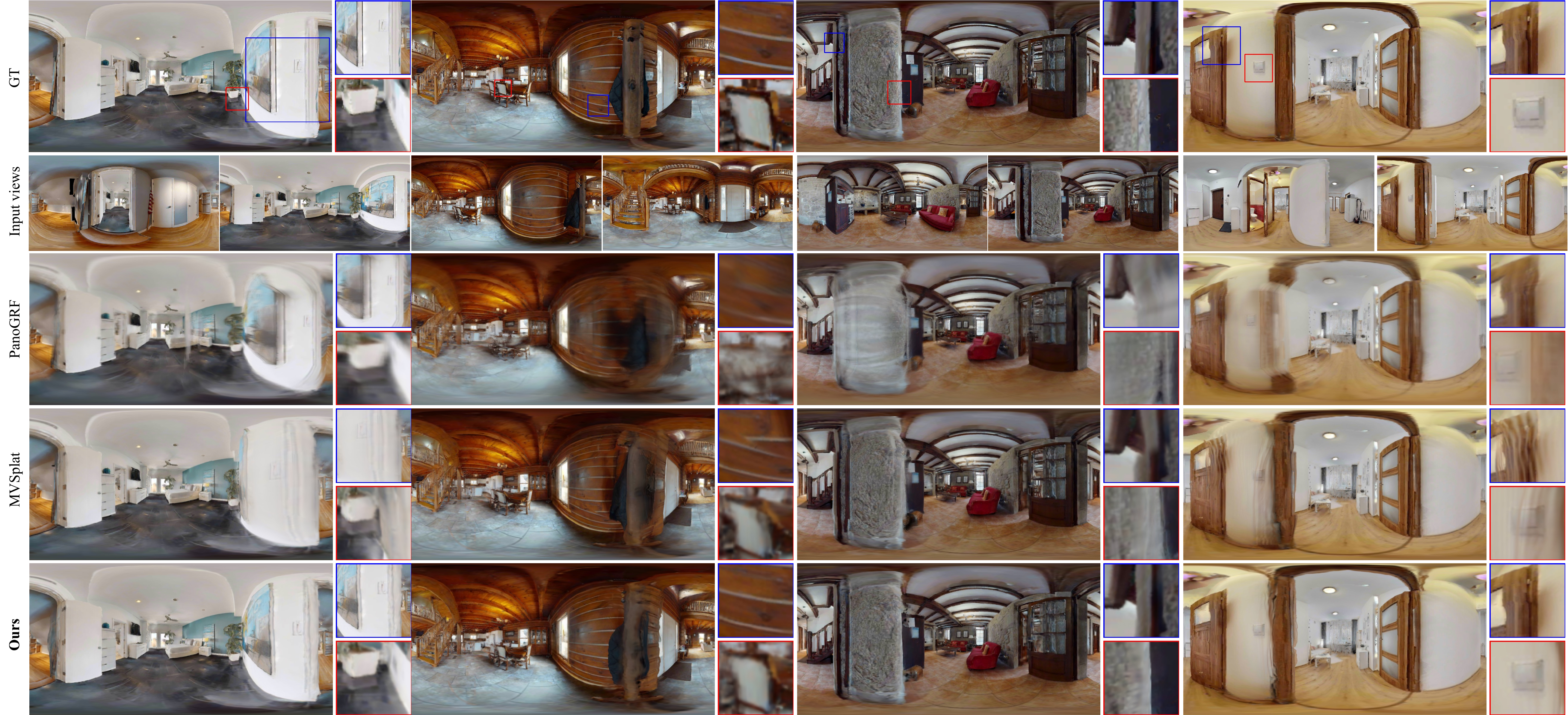}
\caption{Qualitative comparison between our Splatter-360 and PanoGRF, MVSplat on the HM3D dataset. Regions with notable differences are highlighted using red and blue rectangles. Please zoom in for a clearer view.}
\label{fig:hm3d_cmp}
\end{figure*}

\subsection{Implementation Details}

We implement our proposed method using PyTorch and conduct all experiments on a cluster of NVIDIA V100 GPUs, each equipped with 32GB of VRAM. \sysname{} is trained with an RGB loss function that is a linear combination of mean squared error (MSE) and LPIPS loss, with weights set to 1.0 and 0.05, respectively. We also supervise the depth prediction, with the weight set to 0.1. More details about the implementation are presented in the supplementary material.

\subsection{Datasets, Baselines, and Metrics}

\textbf{Datasets.} We evaluate Splatter-360 on two large-scale panoramic datasets: HM3D~\cite{hm3d} and Replica~\cite{replica}. These datasets contain diverse indoor scenes captured with 360° panoramic cameras, providing a challenging benchmark for wide-baseline novel view synthesis.

\vspace{4pt} \noindent \textbf{Baselines.} To proveide a fair comparisions both quantitatively and qualitatively, 
we compare our Splatter-360 with several state-of-the-art generalizable 360-degree method PanoGRF~\cite{panogrf} and generalizable perspective methods including HiSplat \cite{hisplat}, DepthSplat~\cite{depthsplat} and MVSplat \cite{mvsplat}. 
% \chenming{Add}

\vspace{4pt} \noindent \textbf{Metrics.} We evaluate the performance of \sysname{} using standard metrics for novel view synthesis, including PSNR, SSIM, and LPIPS.

% \chenming{I have revised the content above.}

% Additionally, we report the runtime performance of our method to demonstrate its efficiency compared to state-of-the-art methods.
% WE
% \input{image_tex/cmp2}

\begin{figure*}
\centering
\includegraphics[width=\textwidth]{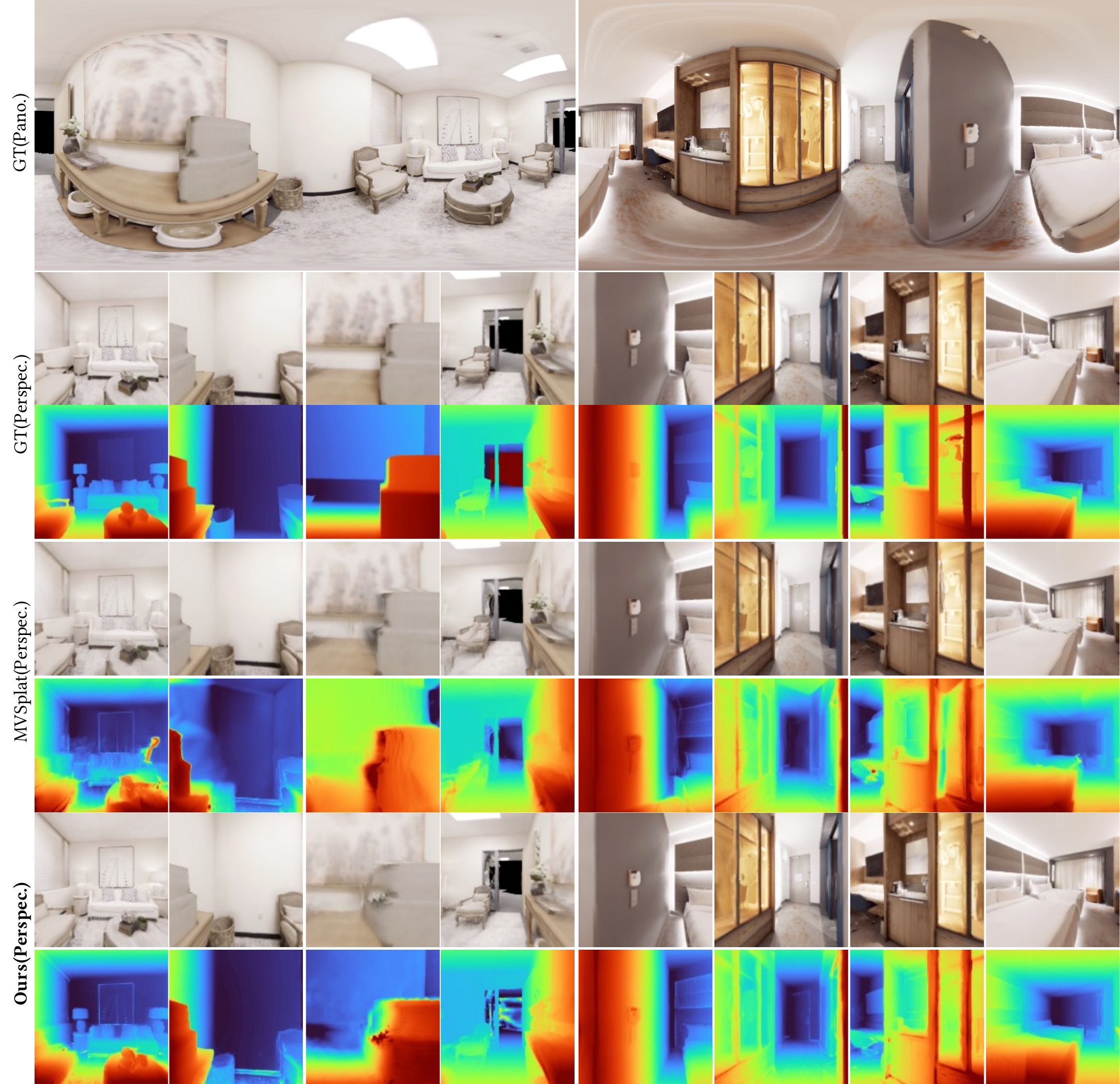}
\caption{Novel view depth comparison between Splatter-360 and PanoGRF on the Replica dataset. ``Pano.'' denotes panoramic view and ``Perspec.'' denotes perspective view. }
\label{fig:replica_depth_cmp}
\end{figure*}

\input{tabs/tab1}

\subsection{Quantitative Results}

Our results demonstrate that \sysname{} outperforms existing methods in both synthesis quality and generalization to novel views. As for the perspective-based method, we convert panoramas to cube maps (12 perspective images) and input these cube maps into HiSplat, DepthSplat, and MVSplat. We did not compare with PixelSplat~\cite{pixelsplat} due to out-of-memory errors during inference with 12 perspective image views. Furthermore, HiSplat and DepthSplat cannot be trained on the HM3D dataset with 12 perspective images due to GPU memory constraints; therefore, we tested their official pre-trained models on Re10K~\cite{re10k} for evaluation on both HM3D and Replica. MVSplat and \sysname{} are trained on HM3D using $8$ Tesla V100 GPUs.

Table~\ref{tab:cmp} shows that existing perspective-based methods, including HiSplat, DepthSplat, and MVSplat, struggle with generalization to cube-map inputs and cannot be directly applied to panoramic datasets. We tested the fine-tuned version of MVSplat, but it still lags behind \sysname{} in terms of generalization, both on HM3D and Replica. Specifically, the PSNR of MVSplat is $1.114$ dB lower than \sysname{} on HM3D and 1.489 dB lower on Replica. We also compared it with PanoGRF~\cite{panogrf}, specifically designed for panoramic inputs. \sysname{} consistently outperforms PanoGRF across all metrics on Replica and HM3D. On HM3D, the PSNR of \sysname{} is $2.662$ dB higher than PanoGRF, and on Replica, it is $1.968$ dB higher.

Moreover, Table~\ref{tab:cmp_depth} provides a quantitative comparison of novel-view depth estimation between MVSplat and \sysname{}. \sysname{} consistently outperforms MVSplat across all depth metrics on HM3D and Replica, verifying that the proposed method can better capture geometry for panoramic inputs.
% rgbd correlation inputs
% \begin{table*}[htbp]
%   \caption{Quantitative comparison with baseline methods on HM3D and Replica Dataset.}
%   \label{tab:cmp_nvs}
%   \setlength\tabcolsep{4pt} % 调小列间距
%   \centering
%   \begin{tabular}{ccccccccc}
%     \toprule
% % \cmidrule(r){1-1}
%     % \multirow{5}{*}{Training data}
%     \multirow{2}{*}{Method} & \multicolumn{3}{c}{HM3D}&\multicolumn{3}{c}{Replica} \\
%         \cmidrule(r){2-4} \cmidrule(r){5-7} 
%    & PSNR$\uparrow$ & SSIM$\uparrow$ & LPIPS$\downarrow$ & PSNR$\uparrow$ & SSIM$\uparrow$ & LPIPS$\downarrow$\\    
%     \cmidrule(r){1-1}    \cmidrule(r){2-4} \cmidrule(r){5-7} 
%     % PixelNeRF~\cite{pixelnerf} & - & - & - & - & - & - \\
%     % MuRF~\cite{murf} & - & - & - & - & - & - \\
%     HiSplat ~\cite{hisplat} & 17.268&0.624	&0.488&  17.157&0.642&0.417\\
%     DepthSplat ~\cite{depthsplat} & 20.224&0.695&0.383&  19.369	&0.732	&0.334  \\
%     MVSplat~\cite{mvsplat} & 27.179&0.851&0.176 & 28.399&0.908&0.115 \\
%    % 27.5954 	0.9082 	0.1129 
%     \sysname{}&{\bfseries28.293 } &{\bfseries0.875}  & {\bfseries0.155} &{\bfseries29.888}&{\bfseries0.924} &{\bfseries0.097 } \\
%     \bottomrule

% % HiSplat and DepthSplat cannot training on HM3D

%   \end{tabular}
% \end{table*}

\subsection{Qualitative Results}
Qualitative comparisons are provided in Fig.~\ref{fig:hm3d_cmp} and Fig.~\ref{fig:replica_cmp}, with key differences highlighted using red and blue rectangles. As shown in the first sample of Fig.~\ref{fig:replica_cmp}, \sysname{} produces much sharper edges and more accurate textures, particularly on objects like the table and chair. In the second and third samples, PanoGRF demonstrates less defined edges, particularly on the floor, while MVSplat exhibits noticeable artifacts in the form of floaters near the white wall. 

A comparison of learned geometry between \sysname{} and MVSplat is presented in Fig.~\ref{fig:replica_depth_cmp}. MVSplat shows significantly poorer depth estimation, as evidenced by its incorrect reconstruction of surfaces such as the lamp and table. In contrast, \sysname{} delivers more accurate and detailed depth predictions with clearer geometry and well-defined surfaces.

% \chenming{Add a few sentences.}

\subsection{Ablation Study}

We validate the effectiveness of different modules proposed in \sysname{} in this section. Due to the large number of ablation studies and limited resources, we utilized $2$ NVIDIA Tesla V100 GPUs for each group of experiments in this section.

\vspace{4pt} \noindent \textbf{Spherical cost volume.}
% Eliminating the spherical cost volume results in a significant drop in performance across all metrics, underscoring its essential role in the model's ability to reconstruct high-quality images. Specifically, as shown in the first row of Table~\ref{tab:ab}, compared to the full model, the PSNR decreases by 5.271 and 2.263 dB on the Replica and HM3D dataset, respectively.
Eliminating the spherical cost volume leads to a notable decline in performance across all metrics, highlighting its crucial role in the model's ability to generate high-quality reconstructions. Specifically, as shown in the first row of Table~\ref{tab:ab}
, compared to the full model, the PSNR drops by $5.271$ dB and $2.263$ dB on the Replica and HM3D datasets, respectively.

\input{tabs/tab3}

\vspace{4pt} \noindent \textbf{ERP encoder vs. CP encoder.}
We separately removed the ERP and CP encoders to evaluate their contributions. The results of these ablations are presented in the third and fourth rows of Table~\ref{tab:ab}. The ERP encoder is crucial for maintaining high PSNR and SSIM values and achieving low LPIPS values, highlighting its vital role in panoramic feature extraction and encoding. In contrast, the removal of the CP encoder results in a smaller performance drop compared to the removal of the ERP encoder. However, removing the CP encoder on the Replica dataset still leads to a PSNR decrease of approximately $0.448$ dB, suggesting that the CP encoder provides auxiliary support in the Gaussian reconstruction of panoramic images.

\vspace{4pt} \noindent \textbf{Cross-view attention.} Corss-view attention is essential to extract the 3D-awareness feature. Here, we test the influence of removing this module. As shown in the second row of Table~\ref{tab:ab}
, such a removing resulted in a noticeable decline in PSNR and SSIM, alongside an increase in LPIPS, indicating the effectiveness of cross-view attention.

\vspace{4pt} \noindent \textbf{Monocular depth network feature.}
Removing the monocular depth encoder led to a PSNR drop of $0.467$ dB in the Replica dataset. This suggests that the pre-trained monocular depth features offer a robust single-view geometry-aware prior, which is especially advantageous for reconstructing 360-degree sparse views in textureless indoor scenes.

\input{tabs/tab2}

%------------------------------------------------------------------------
\section{Conclusion}

This paper introduced Splatter-360, a novel framework for generalizable 3D Gaussian Splatting designed for wide-baseline panoramic images. By leveraging 3D-Aware Bi-Projection Encoding and spherical cost volume, our method significantly improves the quality of novel view synthesis and geometry estimation from panoramic inputs. Experimental results on the HM3D and Replica datasets demonstrate that Splatter-360 sets a new state-of-the-art performance in the panoramic setting, making it a promising solution for applications in VR, simulation rendering, and 360° video streaming.

\vspace{4pt} \noindent \textbf{Limitations and future work.} Our method achieves state-of-the-art performance on several benchmarks but shares some limitations with existing approaches. It requires pose input, lacks generative capabilities, and is currently limited to indoor scenes due to the absence of a large-scale 360$^\circ$ outdoor dataset. In the future, we plan to create a comprehensive synthetic outdoor dataset and enhance our method with a video diffusion model. More importantly, we aim to explore pose-free 360$^\circ$ reconstruction in diverse outdoor environments.

\clearpage

%% file: tabs/tab1.tex
% \begin{table*}[t]
%     \begin{center}
%     \caption{Quantitative comparison with baseline methods on the HM3D and Replica datasets. $^\dagger$ indicates models that were trained by us, whereas for all other methods, we used the pre-trained models provided by the original authors.}
%     \begin{tabular}{lcccccccccccccc}
%     \toprule
%      \multirow{2}{*}[-2pt]{Method} & \multicolumn{3}{c}{HM3D~\cite{hm3d}} & \multicolumn{3}{c}{Replica~\cite{replica}} \\
%     \addlinespace[-12pt] \\
%     \cmidrule(lr){2-4} \cmidrule(lr){5-7} 
%     \addlinespace[-12pt] \\
%      & PSNR$\uparrow$ & SSIM$\uparrow$ & LPIPS$\downarrow$ & PSNR$\uparrow$ & SSIM$\uparrow$ & LPIPS$\downarrow$  \\
%     \midrule
%     HiSplat ~\cite{hisplat} & 17.268&0.624&0.488 
%   & 17.157&0.642&0.417 \\
%      MVSplat ~\cite{mvsplat} & 17.574&0.636	&0.441&  18.005&0.631&0.512 \\
%     DepthSplat ~\cite{depthsplat} & 20.224&0.695&0.383& 19.369	&0.732	&0.334   \\
%     \cmidrule(lr){0-4} \cmidrule(lr){5-7} 
% % 
%     PanoGRF~\cite{panogrf} & 25.631&0.813&0.268  &27.920&0.892&0.171 \\
%     MVSplat$^\dagger$~\cite{mvsplat} & 27.179&0.851&0.176 & 28.399&0.908&0.115  \\
    
%     \sysname{}$^\dagger$&\textbf{28.293} &\textbf{0.875} & \textbf{0.155}  &\textbf{29.888}&\textbf{0.924} &\textbf{0.097} \\
%     \bottomrule
%     \end{tabular}
%     \label{tab:cmp}
%     \end{center}

% \end{table*}
\begin{table*}[t]
    \caption{Quantitative comparison with baseline methods on the HM3D and Replica datasets. $^\dagger$ indicates models that were trained by us on the panoramic dataset, whereas for all other methods, we used the pre-trained models provided by the original authors.}
    \centering
    \setlength{\tabcolsep}{14pt} % 调整列间距，默认是6pt，减小这个值可以让表格更宽
    \begin{tabular}{lccc|ccc}
        \toprule
         & \multicolumn{3}{c}{\textbf{HM3D}~\cite{hm3d}} & \multicolumn{3}{c}{\textbf{Replica}~\cite{replica}} \\
        \cmidrule(lr){2-4} \cmidrule(lr){5-7}
        \textbf{Method}& PSNR$\uparrow$ & SSIM$\uparrow$ & LPIPS$\downarrow$ & PSNR$\uparrow$ & SSIM$\uparrow$ & LPIPS$\downarrow$ \\
        \midrule
        HiSplat~\cite{hisplat} & 17.268 & 0.624 & 0.488 & 17.157 & 0.642 & 0.417 \\
        MVSplat~\cite{mvsplat} & 17.574 & 0.636 & 0.441 & 18.005 & 0.631 & 0.512 \\
        DepthSplat~\cite{depthsplat} & 20.224 & 0.695 & 0.383 & 19.369 & 0.732 & 0.334 \\
        \midrule
        PanoGRF~\cite{panogrf} & 25.631 & 0.813 & 0.268 & 27.920 & 0.892 & 0.171 \\
        MVSplat$^\dagger$~\cite{mvsplat} & 27.179 & 0.851 & 0.176 & 28.399 & 0.908 & 0.115 \\
        Splatter-360$^\dagger$ & \textbf{28.293} & \textbf{0.875} & \textbf{0.155} & \textbf{29.888} & \textbf{0.924} & \textbf{0.097} \\
        \bottomrule
    \end{tabular}
    \label{tab:cmp}
\end{table*}

%% file: tabs/tab3.tex
% \begin{table*}[t]
%     \begin{center}
%     \tablestyle{3.9pt}{1.2}
%     \caption{Quantitative comparision of novel view depth between MVSplat and \sysname{}.}
%     \begin{tabular}{lccccccccccccccc}
%     \toprule
%       Dataset & \multicolumn{4}{c}{Replica~\cite{replica}} & \multicolumn{4}{c}{ HM3D~\cite{hm3d}} \\
%     \addlinespace[-12pt] \\
%     \cmidrule(lr){1-1} \cmidrule(lr){2-5} \cmidrule(lr){6-9} 
%     \addlinespace[-12pt] \\
%      Method & Abs Diff$\downarrow$ & Abs Rel$\downarrow$ & RMSE$\downarrow$ & $\delta < 1.25 \uparrow$& Abs Diff$\downarrow$ & Abs Rel$\downarrow$ & RMSE$\downarrow$ & $\delta < 1.25 \uparrow$ \\
%     \midrule
    
%      % $\times$ SCV  &  23.850&0.818&0.210 & 25.224&0.802&0.223\\    
%      % $\times$ CVA &28.217&0.905&0.124 & 26.918&0.851&0.182\\
%      % $\times$ ERP & 26.985&0.887&0.142  & 25.905&0.827&0.202 \\
%     % & w/o CP MV Transformer & 27.477&\textbf{0.860}&\textbf{0.171} & 29.094&\textbf{0.915}&\textbf{0.110} \\
%     MVSplat & 0.132&0.088&0.247&89.913&0.130&0.094	&0.271&90.469\\
%      \sysname{} &\textbf{0.102}&\textbf{0.063}&\textbf{0.197}&\textbf{94.572}&\textbf{0.106}&\textbf{0.076}	&\textbf{0.223}&\textbf{93.851}\\

%     \bottomrule
%     \end{tabular}
%     \label{tab:cmp_depth}
%     \end{center}
    
% \end{table*}

\begin{table}[t]
    \caption{Estimated depth comparison between MVSplat and Splatter-360 on the Replica and HM3D datasets.}
    \centering
    % \tablestyle{6pt}{1.4}
    \begin{tabular}{lcccc}
        \toprule
        \textbf{Dataset} & \textbf{Metric} & \textbf{MVSplat} & \textbf{Splatter-360} \\
        \midrule
        \multirow{4}{*}{\textbf{Replica}~\cite{replica}} & Abs Diff$\downarrow$ & 0.132 & \textbf{0.102} \\
        & Abs Rel$\downarrow$ & 0.088 & \textbf{0.063} \\
        & RMSE$\downarrow$ & 0.247 & \textbf{0.197} \\
        & $\delta < 1.25$$\uparrow$ & 89.913 & \textbf{94.572} \\
        \midrule
        \multirow{4}{*}{\textbf{HM3D}~\cite{hm3d}} & Abs Diff$\downarrow$ & 0.130 & \textbf{0.106} \\
        & Abs Rel$\downarrow$ & 0.094 & \textbf{0.076} \\
        & RMSE$\downarrow$ & 0.271 & \textbf{0.223} \\
        & $\delta < 1.25$$\uparrow$ & 90.469 & \textbf{93.851} \\
        \bottomrule
    \end{tabular}
    \label{tab:cmp_depth}
\end{table}

%% file: tabs/tab2.tex
\begin{table}[t]
    \begin{center}
    \tablestyle{3.9pt}{1.2}
    \caption{Ablation studies were conducted on the HM3D and Replica datasets. For simplicity, we use the following abbreviations: `SCV' for spherical cost volume and `CVA' for cross-view attention.}
    \begin{tabular}{lccccccccccccccc}
    \toprule
     Ablated & \multicolumn{3}{c}{Replica~\cite{replica}} & \multicolumn{3}{c}{ HM3D~\cite{hm3d}} \\
    \addlinespace[-12pt] \\
    \cmidrule(lr){2-4} \cmidrule(lr){5-7} 
    \addlinespace[-12pt] \\
    module & PSNR$\uparrow$ & SSIM$\uparrow$ & LPIPS$\downarrow$ & PSNR$\uparrow$ & SSIM$\uparrow$ & LPIPS$\downarrow$ \\
    \midrule
    
     $\times$ SCV  &  23.850&0.818&0.210 & 25.224&0.802&0.223\\    
     $\times$ CVA &28.217&0.905&0.124 & 26.918&0.851&0.182\\
     $\times$ ERP & 26.985&0.887&0.142  & 25.905&0.827&0.202 \\
    % & w/o CP MV Transformer & 27.477&\textbf{0.860}&\textbf{0.171} & 29.094&\textbf{0.915}&\textbf{0.110} \\
     $\times$ CP&28.673&0.909&0.117 & 27.277&0.857&0.174\\
     $\times$ Mono Feat.& 28.654&0.911&0.116 & 27.380&0.858&0.173\\
     \midrule 
    \textbf{Full}  & \textbf{29.121} & \textbf{0.914} & \textbf{0.111}  & \textbf{27.487}&\textbf{0.860}&\textbf{0.171}\\
    
    \bottomrule
    \end{tabular}
    \label{tab:ab}
    \end{center}
\end{table}

%% file: sec/X_suppl.tex
\clearpage
\setcounter{page}{1}
\maketitlesupplementary

\section{Additional Quantitative Results}

\subsection{Comparisons with More Input Views}
% Table.\ref{tab:cmp_3views} presents the quantitative comparison of MVSplat and \sysname{} with three view inputs. \sysname{} outperforms MVSplat in SSIM and LPIPS and presents a similar PSNR with MVSplat. We also compared the estimated novel view depth with depth metrics following SimpleRecon~\cite{simplerecon}. \sysname{} significantly outperforms MVSplat in all the depth metrics. \sysname{} shows a good general performance with three view inputs even trained on two view inputs.

Table \ref{tab:cmp_3views} presents a quantitative comparison of MVSplat and \sysname{} using three-view inputs. \sysname{} demonstrates superior performance to MVSplat in SSIM and LPIPS, while exhibiting comparable PSNR values. Additionally, we evaluate the estimated novel view depth using depth metrics used in \cite{simplerecon}. \sysname{} significantly outperforms MVSplat across all the used depth metrics. Our \sysname{} exhibits robust general performance with three-view inputs, despite being trained on two-view inputs.

\begin{table}[b]
    \caption{
    Quantitative comparison with three context views between MVSplat and Splatter-360 on the Replica and HM3D datasets.}
    \centering
    % \tablestyle{6pt}{1.4}
    \begin{tabular}{lcccc}
        \toprule
        \textbf{Dataset} & \textbf{Metric} & \textbf{MVSplat} & \textbf{Splatter-360} \\
        \midrule
        \multirow{7}{*}{\textbf{Replica}~\cite{replica}} &
        PSNR$\uparrow$ & \textbf{29.121}&29.109\\        
        & SSIM$\uparrow$ & 0.908&\textbf{0.913} \\
        & LPIPS$\downarrow$ & 0.123&\textbf{0.116} \\ 
        & Abs Diff$\downarrow$ & 0.125	& \textbf{0.103} \\
        & Abs Rel$\downarrow$ & 0.078&\textbf{0.060} \\
        & RMSE$\downarrow$ & 0.233&\textbf{0.193} \\
        & $\delta < 1.25$$\uparrow$ & 90.771	&\textbf{94.367} \\
        \midrule
        \multirow{7}{*}{\textbf{HM3D}~\cite{hm3d}} 
        & PSNR$\uparrow$ & 27.858&\textbf{27.905}\\
        & SSIM$\uparrow$ & 0.861&\textbf{0.868}\\
        & LPIPS$\downarrow$ & 0.174&\textbf{0.168}\\        
        & Abs Diff$\downarrow$ & 0.118&\textbf{0.095} \\
        & Abs Rel$\downarrow$ &0.083&\textbf{0.067} \\
        & RMSE$\downarrow$ & 0.251&\textbf{0.209} \\
        & $\delta < 1.25$$\uparrow$ & 91.684	&\textbf{94.545} \\
        \bottomrule
    \end{tabular}
    \label{tab:cmp_3views}
\end{table}

\subsection{Comparisons under a Narrow-baseline}
% Table.\ref{tab:cmp_narrow} presents the quantitative comparison of MVSplat and \sysname{} with a narrow baseline. In the main text, we sampled an input pair with a frame interval of 100. Here, the frame interval of the input pair is set to 50. \sysname{} consistently outperforms MVSplat in all the metrics. This indicates that \sysname also has a better general performance under a narrow baseline compared to MVSplat.

Table \ref{tab:cmp_narrow} presents a quantitative comparison between MVSplat and \sysname{} under the narrow-baseline setting. In the main text, we sample an input pair with a frame interval of $100$. Here, the frame interval of the input pair is further reduced to $50$. \sysname{} consistently outperforms MVSplat across all metrics. This result indicates that \sysname{} exhibits superior performance under the narrow-baseline condition.

\begin{table}[b]
    \caption{
    Quantitative comparison under a narrow baseline between MVSplat and Splatter-360 on the Replica and HM3D datasets.}
    \centering
    % \tablestyle{6pt}{1.4}
    \begin{tabular}{lcccc}
        \toprule
        \textbf{Dataset} & \textbf{Metric} & \textbf{MVSplat} & \textbf{Splatter-360} \\
        \midrule
        \multirow{7}{*}{\textbf{Replica}~\cite{replica}} &
        PSNR$\uparrow$ & 32.521&\textbf{33.282}\\        
        & SSIM$\uparrow$ & 0.951&\textbf{0.957}\\
        & LPIPS$\downarrow$ & 0.064&\textbf{0.058} \\ 
        & Abs Diff$\downarrow$ & 0.109&\textbf{0.090} \\
        & Abs Rel$\downarrow$ & 0.057&\textbf{0.048} \\
        & RMSE$\downarrow$ & 0.214&\textbf{0.171} \\
        & $\delta < 1.25$$\uparrow$ & 94.257	&\textbf{96.645} \\
        \midrule
        \multirow{7}{*}{\textbf{HM3D}~\cite{hm3d}} 
        & PSNR$\uparrow$ & 30.851&\textbf{31.493}\\
        & SSIM$\uparrow$ & 0.915&\textbf{0.925}\\
        & LPIPS$\downarrow$ & 0.109&\textbf{0.101}\\
        & Abs Diff$\downarrow$ &  0.102&\textbf{0.092} \\
        & Abs Rel$\downarrow$ &0.060&\textbf{0.058} \\
        & RMSE$\downarrow$ & 0.228&\textbf{0.189} \\
        & $\delta < 1.25$$\uparrow$ & 94.802	&\textbf{96.031} \\
        \bottomrule
    \end{tabular}
    \label{tab:cmp_narrow}
\end{table}

% \subsection{Comparison with PanoGRF training from scratch}

% \subsection{Comparison with MVSplat using depth loss}

% \subsection{Ablation Study of Cost volume Refinement U-Net and Depth Refinement U-UNet}
\subsection{Additional Ablation Studies}
% We conducted additional ablation studies for the cost volume refinement U-Net and depth refinement U-Net. Table~\ref{tab:ab_more} presents the results after we removed these modules respectively. Compared to the full model, the model without depth refinement U-Net PSNR performs much worse. PSNR dropped by about 0.7dB on Replica and by about 0.69dB on HM3D. After we removed cost volume refinement U-Net, PSNR dropped by 0.12dB on Replica and HM3D.

We conduct additional ablation studies on the cost volume refinement U-Net and the depth refinement U-Net. Table~\ref{tab:ab_more} presents the statistical results obtained after removing these modules individually. In comparison to the complete model using all components, the model without depth refinement U-Net exhibits significantly degraded PSNR performance. Specifically, PSNR decreased by approximately 0.7 dB on Replica and by about 0.69 dB on HM3D. Upon removing the cost volume refinement U-Net, PSNR decreased by 0.119 dB on Replica and by 0.125 dB on HM3D.

\begin{table}[t]
    \begin{center}
    \tablestyle{3.9pt}{1.2}
    \caption{Additional ablation studies were conducted on the HM3D and Replica datasets. For simplicity, we use the following abbreviations: `CVRU' for spherical cost volume refinement U-Net and `DRU' for depth refinement U-Net.}
    \begin{tabular}{lccccccccccccccc}
    \toprule
     Ablated & \multicolumn{3}{c}{Replica~\cite{replica}} & \multicolumn{3}{c}{ HM3D~\cite{hm3d}} \\
    \addlinespace[-12pt] \\
    \cmidrule(lr){2-4} \cmidrule(lr){5-7} 
    \addlinespace[-12pt] \\
    module & PSNR$\uparrow$ & SSIM$\uparrow$ & LPIPS$\downarrow$ & PSNR$\uparrow$ & SSIM$\uparrow$ & LPIPS$\downarrow$ \\
    \midrule
    
    %  $\times$ SCV  &  23.850&0.818&0.210 & 25.224&0.802&0.223\\    
    %  $\times$ CVA &28.217&0.905&0.124 & 26.918&0.851&0.182\\
    %  $\times$ ERP & 26.985&0.887&0.142  & 25.905&0.827&0.202 \\
    % % & w/o CP MV Transformer & 27.477&\textbf{0.860}&\textbf{0.171} & 29.094&\textbf{0.915}&\textbf{0.110} \\
    %  $\times$ CP&28.673&0.909&0.117 & 27.277&0.857&0.174\\
     $\times$ DRU & 28.306&0.905&0.129	&26.800	&0.846&0.189\\
     $\times$ CVRU & 29.002&0.912&0.115&27.362	&0.856&0.175\\
     \midrule 
    \textbf{Full}  & \textbf{29.121} & \textbf{0.914} & \textbf{0.111}  & \textbf{27.487}&\textbf{0.860}&\textbf{0.171}\\

    \bottomrule
    \end{tabular}
    \label{tab:ab_more}
    \end{center}
\end{table}

\section{More Implementation Details}
\subsection{Dataset Details}
% \subsubsection{Dataset Building}
The datasets are built based on Replica~\cite{replica} and HM3D~\cite{hm3d} textured mesh dataset. In particular, we sample camera trajectories to render videos with AI-Habitat simulation tool~\cite{habitat}. Since AI-habitat only provides the API for capturing perspective views, we first get cube maps for each viewpoint and stitch them into panoramas.

% 数据集训练测试集划分
For HM3D~\cite{hm3d}, we split the train and test set following their original split. HM3D contains 800 training scenes and 100 test scenes. We sample 5 camera trajectories for each scene. In total, we finally rendered 4000 training scenes and 500 test scenes.

For Replica, we use all the scenes for testing.
Replica has 18 scenes in total, and we sample 5 camera trajectories for each scene. In total, we render 90 test scenes.
We randomly sample 3 target views between the context image pair for testing.

\subsection{Experiment Details}
In the comparisons of the main paper, for HiSplat, MVSplat, and DepthSplat, we utilize their model pre-trained models on RE10K~\cite{re10k} for evaluation. We set $near=0.5$ and $far=10$ for these models as these parameters are relatively close to their training setting. We match features under the resolution of $\frac{1}{4}H\times\frac{1}{4}W$, where $H$ and $W$ are the height and width of input images.  We apply $near=0.1$ for HiSplat, MVSplat, and DepthSplat, but the results get worse as $near=0.1$ is much different from their training setting on RE10K~\cite{re10k}.

For MVSplat$^\dagger$ and \sysname{}$^\dagger$ trained on HM3D, we set $near=0.1$ and $far=10$ to match our indoor dataset for the consideration of fairness. We perform cross view matching under the resolution of $\frac{1}{8}H\times\frac{1}{8}W$ due to GPU memory limits.

\subsection{Network Details}
% Multi-view Transformer
We adopt the encoder of UniMatch~\cite{unimatch} as our backbone. The first convolution layer downsamples images with a stride of 2. Next, we utilize six residual layers to extract features. The first two residual layers contain utilize the stride of 1. Subsequently, we downsample features in half after every two residual layers with a stride of 2. We then get $\frac{1}{8}H\times\frac{1}{8}W$ feature maps. The downsampled feature maps are fed into a cross-view transformer composed of six stacked transformer blocks. Each transformer block contains a self-attention and a cross-view attention layer. Similar to MVSplat~\cite{mvsplat}, we utilize the local window attention of SwinTransformer~\cite{swintransformer}.  We apply the network architecture for our ERP multi-view transformer and CP multi-view transformer.

For the cost volume refinement U-UNet, we adopt the U-Net from Stable Diffusion 1.5~\cite{sd} as our implementation with an unchanged feature channel of 128 throughout the network. We apply two times $2\times$ down-sampling and one self-attention layer at the $4\times$ down-sampled level. We flatten the feature map before feeding them to the attention module, to interact with the features among different views utilizing the multi-view attention similar to ~\cite{mvdream}. For the depth refinement U-Net which we omitted in the main text for simplicity, we apply 4 times $2\times$ down-sampling and add the multi-view attention at $16\times$ down-sampled level.

We set $D=128$ in the depth sampling consistently with MVSplat~\cite{mvsplat}

\section{Preliminary of 3DGS}
The 3D Gaussian ellipsoid is formally defined as:
\begin{equation}
    G(\boldsymbol{x} \mid \boldsymbol{\mu}, \boldsymbol{\Sigma}) = e^{-\frac{1}{2}(\boldsymbol{x}-\boldsymbol{\mu})^T \boldsymbol{\Sigma}^{-1}(\boldsymbol{x}-\boldsymbol{\mu})}
\end{equation}
where $\boldsymbol{\mu} \in \mathbb{R}^{3}$ represents the spatial mean, and $\boldsymbol{\Sigma} \in \mathbb{R}^{3 \times 3}$ denotes the covariance matrix. To ensure numerical stability during optimization, the covariance matrix $\boldsymbol{\Sigma}$ is decomposed into a scaling matrix $\boldsymbol{S}$ and a rotation matrix $\boldsymbol{R}$ as follows:
\begin{equation}
    \boldsymbol{\Sigma} = \boldsymbol{R} \boldsymbol{S} \boldsymbol{S}^{\top} \boldsymbol{R}^{\top}
\end{equation}

During the rendering process, the 3D Gaussians are projected onto a 2D image plane. Using the intrinsic matrix $\boldsymbol{K}$ and extrinsic matrix $\boldsymbol{T}$, the 2D mean $\boldsymbol{\mu}'$ and covariance matrix $\boldsymbol{\Sigma}'$ are computed as:
\begin{equation}
    \boldsymbol{\mu}^{\prime} = \boldsymbol{K}[\boldsymbol{\mu}, 1]^{\top}, \quad \boldsymbol{\Sigma}^{\prime} = \boldsymbol{J} \boldsymbol{T} \boldsymbol{\Sigma} \boldsymbol{T}^{\top} \boldsymbol{J}^{\top}
\end{equation}
Here, $\boldsymbol{J}$ represents the Jacobian matrix of the affine approximation of the projective transformation. Each Gaussian is associated with an opacity value $o$ and a view-dependent color $\boldsymbol{c}$, which is determined by a set of spherical harmonics coefficients. The pixel color $\boldsymbol{C}$ is computed via alpha-blending over the 2D Gaussians, sorted from front to back:
\begin{equation}
    \boldsymbol{C} = \sum_{i \in N} T_i G_i\left(\boldsymbol{u} \mid \boldsymbol{\mu}^{\prime}, \boldsymbol{\Sigma}^{\prime}\right) \sigma_i \boldsymbol{c}_i
\end{equation}
where the transmittance $T_i$ is defined as:
\begin{equation}
    T_i = \prod_{j=1}^{i-1}\left(1 - G_i\left(\boldsymbol{u} \mid \boldsymbol{\mu}^{\prime}, \boldsymbol{\Sigma}^{\prime}\right) \sigma_i \right)
\end{equation}